\documentclass{article}

\usepackage[english]{babel}
\PassOptionsToPackage{numbers, compress}{natbib}
\usepackage[preprint]{neurips_2023}

\usepackage{caption}
\usepackage{amsmath}
\usepackage{graphicx}
\usepackage{amsfonts}
\usepackage{subfigure}

\usepackage[utf8]{inputenc} 
\usepackage[T1]{fontenc}    
\usepackage{hyperref}       
\usepackage{url}            
\usepackage{booktabs}       
\usepackage{amsfonts}       
\usepackage{nicefrac}       
\usepackage{microtype}      
\usepackage{xcolor}         

\title{\textbf{Text-guided High-definition Consistency Texture Model}}
\author{Zhibin Tang \\
Midea AIIC\\
\texttt{tangzb14@midea.com}
\And Tiantong He\\
Midea AIIC\\
\texttt{hhlovesriya@gmail.com}}

\begin{document}
\maketitle

\begin{abstract}
With the advent of depth-to-image diffusion models, text-guided generation, editing, and transfer of realistic textures are no longer difficult. However, due to the limitations of pre-trained diffusion models, they can only create low-resolution, inconsistent textures. To address this issue, we present the High-definition Consistency Texture Model (HCTM), a novel method that can generate high-definition and consistent  textures for 3D meshes according to the text prompts. We achieve this by leveraging a pre-trained depth-to-image diffusion model to generate single viewpoint results based on the text prompt and a depth map. We fine-tune the diffusion model with Parameter-Efficient Fine-Tuning to quickly learn the style of the generated result, and apply the multi-diffusion strategy to produce high-resolution and consistent results from different viewpoints. Furthermore, we propose a strategy that prevents the appearance of noise on the textures caused by backpropagation. Our proposed approach has demonstrated promising results in generating high-definition and consistent textures for 3D meshes, as demonstrated through a series of experiments.
\end{abstract}

\section{Introduction}
In recent years, the field of artificial intelligence has experienced a significant resurgence, driven in large part by advances in machine learning techniques, particularly generative models. These models have become increasingly popular in a variety of fields, including computer vision, natural language processing, and speech recognition.

Text-to-image generative models are a disruptive technology, which make synthesizing high-quality and diverse images from text prompts become a reality. Given a textual description, these new models demonstrate unprecedented capabilities in generating highly detailed imagery that captures the essence and intent of the input text. Despite the breakthrough in text-to-image generation, generating high quality 3D models remains a significant challenge.

Recent research has made significant progress in the field of painting and texturing 3D objects using 2D vision-language prior knowledge. Magic3D\cite{lin2023magic3d}, Latent- NeRF\cite{metzer2022latent} apply score distillation\cite{poole2022dreamfusion} to indirectly utilize Stable Diffusion\cite{rombach2022high} as a texturing prior. TEXTure\cite{richardson2023texture} employs a denoising process on rendered images by utilizing a pre-trained depth-conditioned diffusion model.

Despite the impressive quality achieved by these methods, they still fall short in comparison to their 2D counterparts in terms of quality. We attribute this to the inconsistency between different viewpoints during rendering and the low-resolution of the generated images. To be more precise, their postulation is that neural networks that have been trained can produce textures on meshes that are realistic enough to make the resulting 3D objects appear genuine at first glance. However, such assumptions may not always hold true in real-world scenarios.

 This paper introduces HCTM, a straightforward but highly effective texture synthesis method that ensures viewpoint consistency and high-definition by utilizing the diffusion model. Our motivation is straightforward: no single model is capable of capturing all possible viewpoints of objects, but a model that adapts to the data at hand may be the solution to this problem. Our methodology involves heavy data augmentation, where object-irrelevant features such as noisy backgrounds are filtered during training. To ensure the transferred style are consistent, we use Parameter-Efficient Fine-Tuning algorithm to let the diffusion learn the knowledge of the target data style quickly. Then we apply multi-diffusion strategy on our fine-tuned diffusion model to generate high-resolution and consistent results from different views. Besides, we introduce a strategy that prevents the appearance of the noise on the textures caused by backpropagation.

We evaluate HCTM and find it to be highly effective for texture generation and style transfer. More importantly, our evaluations show that HCTM produces textures of significantly higher quality than previous methods. In terms of computational cost, our method only takes minimal resources during training and negligible computational overhead during testing. We believe that HCTM offers a novel perspective on texture synthesis.

\section{Related Works}

\subsection{Diffusion Model}
Diffusion models\cite{ho2020denoising,song2022denoising} are a class of generative models that have gained significant attention recently due to their impressive performance in image and video synthesis tasks. These models are based on the theoretical foundation of asymptotic analysis of noise cluster distribution, and employ dynamic adjustment of the diffusion step size to ensure algorithmic stability.

Diffusion models have shown remarkable performance in various domains, including images\cite{dhariwal2021diffusion,yang2022paint}, videos\cite{ho2022imagen,singer2022makeavideo}, 3D scenes\cite{müller2023diffrf,tang2023make}, and motion sequences\cite{tevet2022human,yuan2022physdiff}. In the field of text-to-image generation, diffusion models have demonstrated impressive generation effects, especially the Stable Diffusion model\cite{rombach2022high}. This model is trained on a large amount of text-image dataset, utilizes CLIP\cite{radford2021learning} to encode text prompts, and uses VQ-VAE\cite{oord2018neural} to encode images into latent spaces, achieving high-quality text-to-image generation.
\subsection{Controllable generation with diffusion models}
Although diffusion models have shown remarkable performance in image generation tasks, their controllability remains a major challenge. Controlling the denoising process in diffusion models is difficult, which limits their practical applications. To address this issue, researchers have proposed various methods to enhance the controllability of diffusion models.

These methods can be broadly classified into two categories. The first category involves incorporating explicit control by using additional guiding signals to the model\cite{avrahami2023spatext,rombach2022highresolution,brooks2023instructpix2pix} . These guiding signals include spatial and textual guidance, high-resolution training datasets, and instruction-based methods. However, these methods require extensive training on curated datasets, which can be time-consuming and expensive. ControlNet\cite{zhang2023adding} incorporates additional information such as depth, segmentation and sketching into the diffusion as conditions. By using zero convolution fine-tuning diffusion, the generation of diffusion can be controlled under certain conditions. 

The second category involves implicitly controlling the generated content by manipulating the generation process of a pre-trained model\cite{kwon2023diffusionbased,meng2022sdedit,tumanyan2022plugandplay} or performing lightweight model fine-tuning\cite{ruiz2023dreambooth,kawar2023imagic,kim2022diffusionclip}. These methods are typically more efficient and require less training data than explicit control methods. For instance, MultiDiffusion\cite{bartal2023multidiffusion} enhances the continuity of super-resolution image generation by reconciling denoising sampling step of different regions.
\subsection{3D Texture Generation}
While Stable Diffusion has shown impressive results in 2D image generation, generating 3D textures is still a challenging task due to its diversity and complexity. To overcome this challenge, researchers have proposed several methods that aim to generate high-quality 3D textures. GET3D\cite{gao2022get3d} queries the texture field at surface points to get colors and use a rasterization-based differentiable renderer to obtain RGB images and silhouettes. Nvdiffrec\cite{Munkberg_2022_CVPR} jointly learns the topology, materials, and environment map lighting from 2D supervision. The authors employ a differentiable variant of the split sum approximation method in order to effectively address environment lighting in their model. CLIP-Mesh\cite{Mohammad_Khalid_2022} employs CLIP-space similarity measurements as an optimization objective, engendering the generation of innovative geometries and textures. Latent-NeRF\cite{metzer2022latent} demonstrates the applicability of score distillation loss within the latent space of Stable Diffusion for generating latent 3D NeRF models. Furthermore, the authors introduce a novel texture generation approach, termed Latent-Paint, which synthesizes high-quality textures by rendering latent texture maps using score distillation and subsequently decoding them into RGB format for the ultimate colorization output. TEXTure\cite{richardson2023texture} iteratively renders the object from different viewpoints, applies a depth-based painting scheme, and projects it back to an atlas.

\subsection{Parameter-Efficient Fine-Tuning}
The Stable Diffusion model has a large number of parameters and requires a significant amount of training data, making it difficult to adjust the model without retraining the entire diffusion parameters. Recent research has focused on developing Parameter-Efficient Fine-Tuning (PEFT) methods to enable the customization and personalization of text-to-image diffusion models with only a few personalized images. 
PEFT methods can fine-tune different parts of the model, including the text embedding, full weights, or cross-attention layers. 

Existing adapter modules have bottleneck serial architecture and can be inserted into every Transformer layer. For instance, LoRA \cite{hu2021lora} assumes low-rank intrinsic dimensionality and performs low-rank updates, while Prefix-Tuning\cite{roich2022pivotal} and prompt-tuning methods append a learnable vector to the attention heads at each Transformer layer or only to the input embedding. UniPELT \cite{mao2022unipelt} integrates multiple PEFT modules with a dynamic gating mechanism, while AdaMix \cite{wang2022adamix} leverages weight averaging for a mixture of adapters.

\section{Method }

\subsection{Motivation and overview}
The diffusion-based text-to-3D methodologies leverage the depth-to-image functionality of diffusion processes to generate realistic textures. Although these methods can produce visually appealing results, the resolution and consistency of the textures are constrained by the limitations of the pretrained diffusion models. Prior works have attempted to address the consistency issue by dynamically defining a trimap partitioning of the rendered image into three progressive states. However, this approach merely refines additional regions without significantly improving the overall consistency.

\begin{figure*}[h]
\centering
\includegraphics[width=0.8\textwidth]{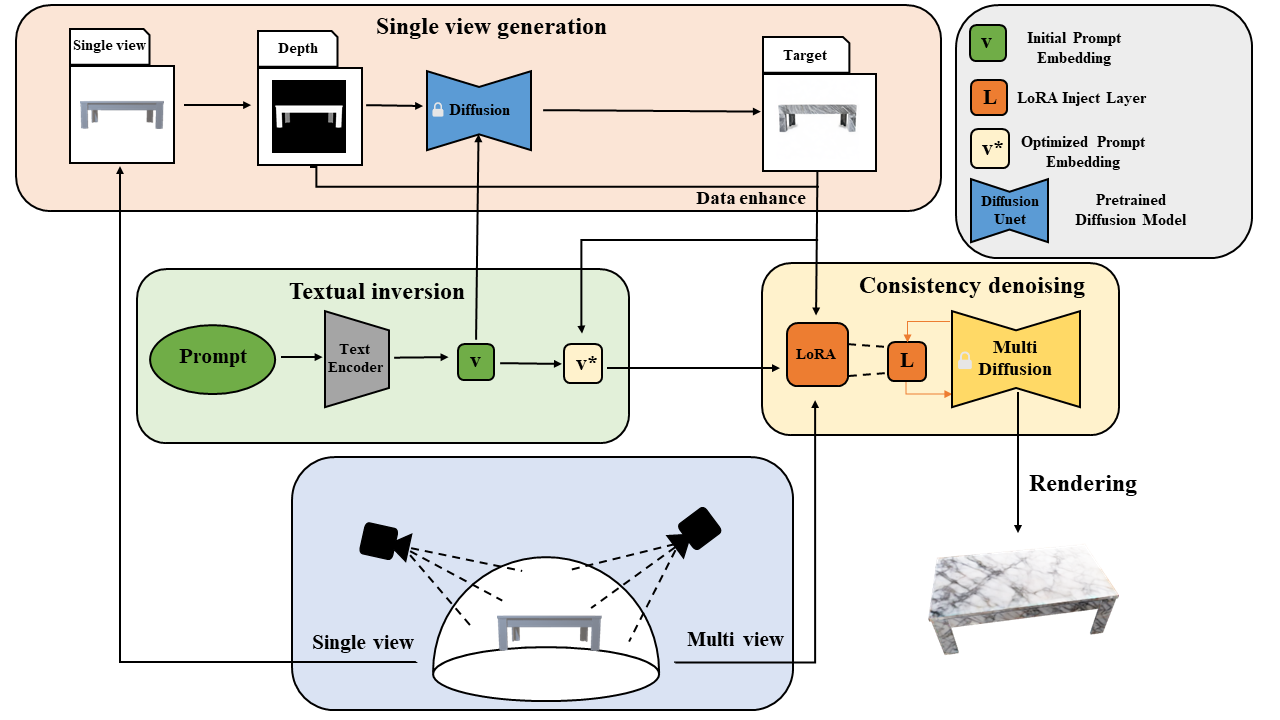}
\caption{\label{fig:1}Given a mesh and a prompt(e.g."marble dining table" in this figure), choose a camera pose $\pi$ to obtain the depth map for this view using pre-trained diffusion model to generate a target image. Optimize the text prompt embedding and  diffusion model weight by textual inversion and LoRA technology based on the target image after data enhancement. Apply multi-diffusion strategy to get the high-definition consistency texture for each viewpoint, and backpropagate the corresponding texture to get the final rendering result.}
\end{figure*}

To overcome these limitations, we propose HCTM, a novel framework for high-definition and consistent texture generation. In the subsequent sections, we provide a comprehensive overview of each component of HCTM. The overall architecture of our framework is depicted in Figure~\ref{fig:1} and discussed in detail using academic language.

\subsection{Single view generation}
The single view generation component of our HCTM framework involves the following steps. Firstly, we randomly select a camera pose $\pi$ for a given 3D mesh, and then obtain the corresponding normalized depth map $d$. This depth map is used as input for a pre-trained depth-to-image diffusion model, which generates a multitude of images that are consistent with the depth map. Out of these generated images, we select one as the target image for further processing. The selection of the target image can be done based on various criteria such as visual quality, clipping scores, or other deep learning image quality assessment techniques.

Furthermore, we also explore the reverse process of this step. By sourcing images from the internet, we can predict the depth maps of these images using depth prediction models such as MiDaS \cite{ranftl2020robust}. This enables us to use a wider range of real-world images as input for our framework, expanding the scope of texture generation to a more diverse range of objects and scenes.

To address the issue of limited training samples, we employ data augmentation techniques such as rotation, translation, scaling, flipping, and background replacement. These techniques ensure that the model can generalize well to unseen viewpoints.

\subsection{Textual inversion}
The issue of inconsistency across different viewpoints arises due to the broad and ambiguous nature of text prompts. Describing an object based on a single prompt can be highly ambiguous. In computer graphics, the Spatially Varying Bidirectional Reflectance Distribution Function (SVBRDF) is employed to characterize the physical properties of an object's surface. However, the same text prompt can correspond to remarkably distinct SVBRDF properties. For instance, a detailed description of a marble table may still correspond to varying SVBRDF values, complicating the consistency of SVBRDF and consequently affecting the rendering process.

To ensure SVBRDF consistency, we employ image inversion to associate the text prompt with an image. To optimize our text prompt, we utilize textual inversion\cite{gal2022image}, which enables the acquisition of a new text prompt embedding $v^*$. This new text prompt embedding $v^*$ better aligns with the target image, as determined by the minimization function:
\begin{align}
v^*=\arg\min_{v}\mathbb{E}_{z\sim\mathcal{E}(x),y,\epsilon\sim\mathcal{N}(0,1),t,d}\left[\|\epsilon-\epsilon_\theta(z_t,t,d,c_\theta(y))\|_2^2\right]
\end{align}
Let $c_\theta$ represent the text encoder that encodes a text prompt input $y$ into a vector in latent space. $t$ denotes the time step, $z_t$ refers to the latent noise at time $t$, $\epsilon$ signifies the unscaled noise sample, and $\epsilon_\theta$ represents the denoising network. During training, $c_\theta$ and $\epsilon_\theta$ are jointly optimized to minimize the given function. At the inference stage, a random noise tensor is sampled and iteratively denoised to generate a new image latent, $z_0$.

\subsection{Consistency denoising }
Although optimized text prompt embedding better describes the target image, many details of the target image are still difficult to describe by text prompt embedding. Parameter-Efficient Fine-Tuning is undoubtedly the most effective method to overcome this problem, which allows the diffusion model to quickly learn the feature of the details that cannot be described by text prompt embedding.
We adopt LoRA technique motivated by \cite{lora}.
For the pretrained weight $W\in\mathbb{R}^{d\times k}$ of Stable Diffusion,instead of updating the entire model,LoRA constrain the update by representing the latter with a low-rank decomposition, and the rank $ r\ll  min(d, k)$,we set $r=32$ in our experiment.
\begin{align}
W+\Delta W=W+BA,~B\in\mathbb{R}^{d\times r},~A\in\mathbb{R}^{r\times k}
\end{align}
Utilizing Parameter-Efficient Fine-Tuning technique, such like LoRA, has demonstrated the capacity to address consistency concerns, albeit without enhancing the resolution of the diffusion output. In order to obtain higher resolution images, we adopt the multi-diffusion strategy.
For a potentially image space
$\mathcal{J}\in\mathbb{R}^{H^{\prime}\times W^{\prime}\times C}$
starts with some initial noisy input $J_{T}\sim N(0,1)$  .With the condition $v^*$,$d$, the multi-diffusion denoising step $\Psi $ is the solution of the optimization equation
\begin{align}
\Psi(J_t|v^*,d)=\underset{J\in\mathcal{J}}{\arg\min}~\mathcal{L}_{\mathrm{MD}}(J|J_t,v^*,d)
\end{align}
\begin{align}
\mathcal{L}_{\mathrm{MD}}(J|J_t,v^*,d)=
\sum_{i=1}^{n}\left\|W_i\otimes\left[F_i(J)-\Phi(I_t^i|v^*,d_i)\right]\right\|^2
\end{align}
where $W_i\:\in\:\mathbb{R}^{H\times W}_+$  are per pixel weights and $\otimes $ is the Hadamard product.
The MD loss reconciles the different denoising sampling steps $\Phi$  suggested on different regions of the generated image $J_t$. Regions $F_i$ is choosed by sliding window in the latent space, and $H=W=64$ defined in the Stable Diffusion. Meanwhile, we remove the regions that don’t intersect with the update mask defined by TEXTure\cite{richardson2023texture}. Furthermore, we set $W_i$ as the proportion of object mask in $F_i$ to reduce the influence of the background on the entire multi-diffusion process.

\subsection{Rendering and texture projection}
To project back diffusion output $I_t$ to the texture ${\cal T}_{t}$ ,we apply gradient-based optimization for ${\cal L}_{t}$ over the values of ${\cal T}_{t}$ when rendered through the differential renderer $\cal R$. That is
\begin{align}
{\cal L}_{t}=\left\|{\cal R}(V,\mathbf{n},\pi,{\cal T}_{t})-I_{t}\right\|_2^2\odot m_{s}
\end{align}
Where $V$ is the vertex property of the mesh, $\mathbf{n}$ is the normal of the mesh, ${\cal T}_{t}$ is the texture, and $\pi$ represents the information associated with the scene observed from this particular viewpoint.

Because we only care about the rendering results of individual objects and do not consider the overall lighting effect on object rendering. We use the local illumination rendering method. We can describe the rendering process using the following formula.
\begin{align}
{\cal R}=I_l{\cal T}_{t}+I_s
\end{align}
We will introduce two rendering methods, the Cook-Torrance microfacet specular shading model and the Spherical Harmonic model, to address different rendering needs in different scenarios.

\textbf{The Cook-Torrance microfacet specular shading model:} 

For the rendering of the diffuse part, we aim to minimize computational cost, so we use the most basic Phong diffuse model:
\begin{align}
I_l=k_d(\mathbf{l}\cdot\mathbf{n}),
\end{align}
where, $k_d$ is diffuse term. $\mathbf{l}$ , $\mathbf{n}$ , are light direction, normal vector.

For the rendering of the specular part, we use the following formula for calculation:
\begin{align}
I_s=\dfrac{D(\mathbf{h})F(\mathbf{v},\mathbf{h})G(\mathbf{l},\mathbf{v},\mathbf{h})}{4(\mathbf{n}\cdot\mathbf{l})(\mathbf{n}\cdot\mathbf{v})}
\end{align}
where $\mathbf{l}$ and $\mathbf{n}$ denote the light and normal vectors, respectively. $\mathbf{h}$ signifies the half-angle direction vector, $\mathbf{v}$ represents the halfway vector, and the vectors $\mathbf{l}$, $\mathbf{h}$, and $\mathbf{v}$ can all be derived from the scene information $\pi$.
In this formula, the term $D$ denotes the normal distribution function, and we employ Disney's selection of the GGX/Trowbridge-Reitz model for its computation due to its lower computational costs. The term $F$ corresponds to the Fresnel term, for which we utilize Schlick's approximation. The term $G$ signifies the specular geometric attenuation term, and we apply the Smith model for GGX in its calculation. For further details, kindly refer to the cited source, as we will not delve into additional specifics here\cite{karis2013real}.

Ultimately, our rendering equation $\cal R$ is the sum of the diffuse and specular part.
\begin{align}
{\cal R}={\cal T}_{t}k_d(\mathbf{l}\cdot\mathbf{n})+\dfrac{D(\mathbf{h})F(\mathbf{v},\mathbf{h})G(\mathbf{l},\mathbf{v},\mathbf{h})}{4(\mathbf{n}\cdot\mathbf{l})(\mathbf{n}\cdot\mathbf{v})}
\end{align}
\textbf{The Spherical Harmonic model:}
\begin{align}
{\cal R}={\cal T}_{t}\sum\limits_{l=0}^{n-1}\sum\limits_{m=-l}^l w_l^m Y_l^m(\mathbf{n}),\quad
\end{align}
Here, $Y_l$ is determined by normals $\mathbf{n}$ while $I_s$ is set to 0.
The $Y_l^m$ represents an orthonormal basis for spherical functions, which is analogous to a Fourier series. In this representation, $l$ denotes the frequency, and $w_l^m$ signifies the corresponding coefficient for a specific basis function. Specifically, by setting the value of $l$ to 3, a total of 9 coefficients are predicted. By manipulating various $Y_l^m$ components, different lighting effects can be simulated, as described by \cite{10.1145/383259.383317}.

In scenarios where the direction of light is provided, the Cook-Torrance microfacet specular shading model is employed for rendering purposes. The SH(Spherical Harmonic) model is utilized in the study to prevent specular reflections from influencing the backpropagation results. SH offers a more uniform diffusion and mitigates the impact of specular reflections.

This method can project back the result without reverse uv mapping, but it is also the main cause of a lot of noise at the border of the mask. This is because in uv mapping, we use interpolation which leads to insufficient constraints on the pixels on the border of the mask. For instance: 
\begin{align}
P=\sum\limits_{i=1}^{4} w_iP_i ~~~~~~~s.t.~~ w_1+w_2+w_3+w_4=1
\end{align}
 $P_1, P_2, P_3, P_4$ is the pixels on the texture, $P$ is the pixel value on render result. Obviously, only know the value of $P$, can not solve the value of $P_1, P_2, P_3, P_4$, since the number of unknowns is greater than the number of equation. Therefore, we need to add some constraints to limit the solution space. Here, we constrain the value of the derivative of each pixel
\begin{align}
{\cal L}_{t}=(\left\|{\cal R}(V,\mathbf{n},\pi,{\cal T}_{t})-I_{t}\right\|_2^2+\lambda\left |\bigtriangledown {\cal R}\right |)\odot m_{s}
\end{align}
We set $\lambda=0.01$ in our experiment. For the previous example, there is a unique optimal solution, $P_1=P_2=P_3=P_4=P$.
\subsection{More detail}
Our texture is represented as a 2048 × 2048 atlas, where the rendering resolution is 2400 × 2400. For the multi-diffusion, our output resolution is 1024 × 1024, stride is 16. All shapes are rendered with 8 viewpoints around the object, and two additional top/bottom views. HCTM is trained on a single Nvidia RTX 3090 GPU and takes about 15 minutes for LoRA technique, 20 minutes for texture generation.

\section{Experiment }
In the following section, we compare our method with two state-of-the-art methods, Latent-NeRF\cite{metzer2022latent} and TEXTure\cite{richardson2023texture}. We compare three different materials separately on the same simple mesh to demonstrate the advantages of our method in terms of consistency, clarity and stability. For complex mesh, we show the visual superiority of our generated texture through user study. Besides, we demonstrate the capability of our method on style transfer.

\subsection{Consistency}
\begin{figure}[h]
\centering
\subfigure{
\rotatebox{90}{\normalsize{~~Latent-NeRF}}
\begin{minipage}[t]{0.75\columnwidth}
\centering
\includegraphics[width=0.75\columnwidth]{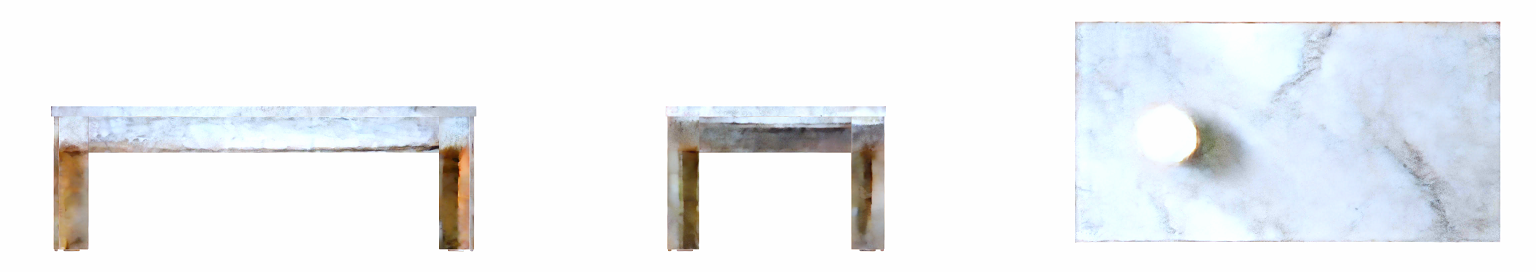}  
\end{minipage}
}
\subfigure{
\rotatebox{90}{\normalsize{~~~TEXTure}}
\begin{minipage}[t]{0.75\columnwidth}
\centering
\includegraphics[width=0.75\columnwidth]{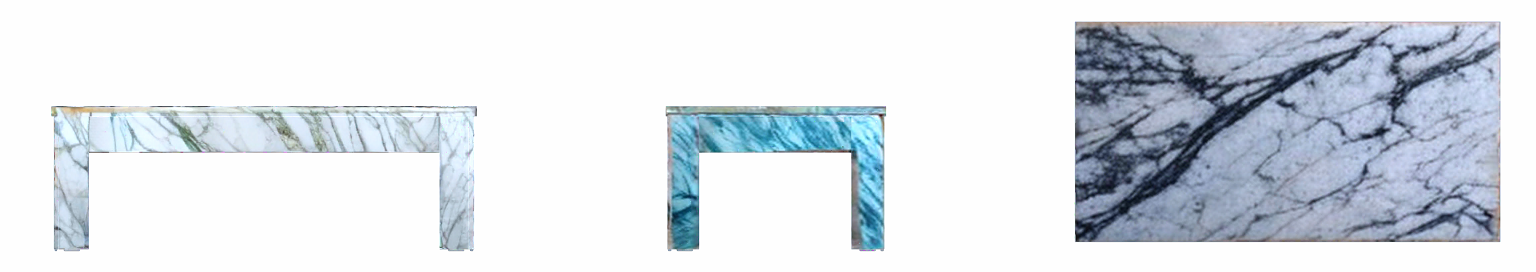}   
\end{minipage}
}
\subfigure{
\rotatebox{90}{\normalsize{~~~~Ours}}
\begin{minipage}[t]{0.75\columnwidth}
\centering
\includegraphics[width=0.75\columnwidth]{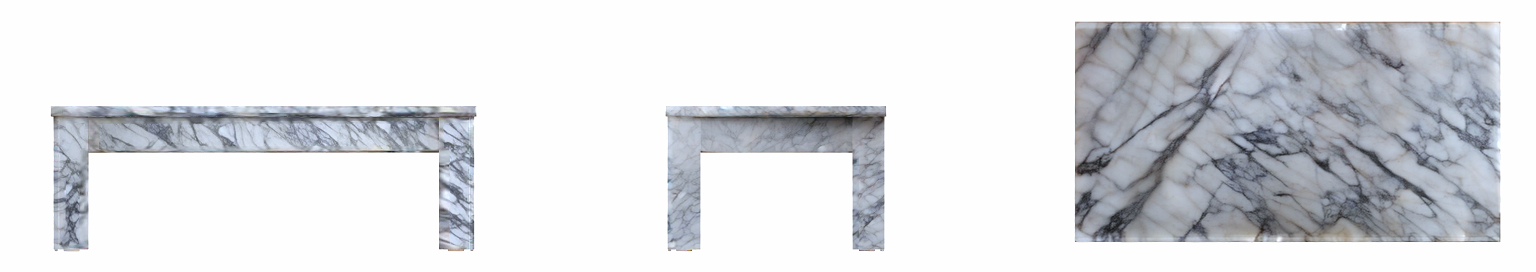}   
\end{minipage}
}
\caption{\label{fig:2}3D generated results of Latent-NeRF, TEXTure, and HCTM given the same mesh and text prompt "marble dining table". HCTM produces textures of higher consistency.}
\end{figure}

To evaluate the consistency of our method, we conduct experiments comparing our approach with two state-of-the-art methods using the text prompt "marble dining table". In Figure~\ref{fig:2}, we present the generated textures from each method for different viewpoints. It can be observed that while Latent-NeRF only barely resembles marble from the top view, and other perspectives are far from the desired material. The generated textures from TEXTure are indeed marble, but the color and pattern from each perspective are completely different. On the contrary, our method maintains a high degree of consistency in both color and pattern across all viewpoints. This indicates that our method is capable of generating high-quality textures with consistent visual characteristics, even for complex materials like marble.

\subsection{Clarity}
\begin{figure}[h]
\centering
\subfigure{
\rotatebox{90}{\normalsize{~~~~Latent-NeRF}}
\begin{minipage}[t]{0.75\columnwidth}
\centering
\includegraphics[width=0.75\columnwidth]{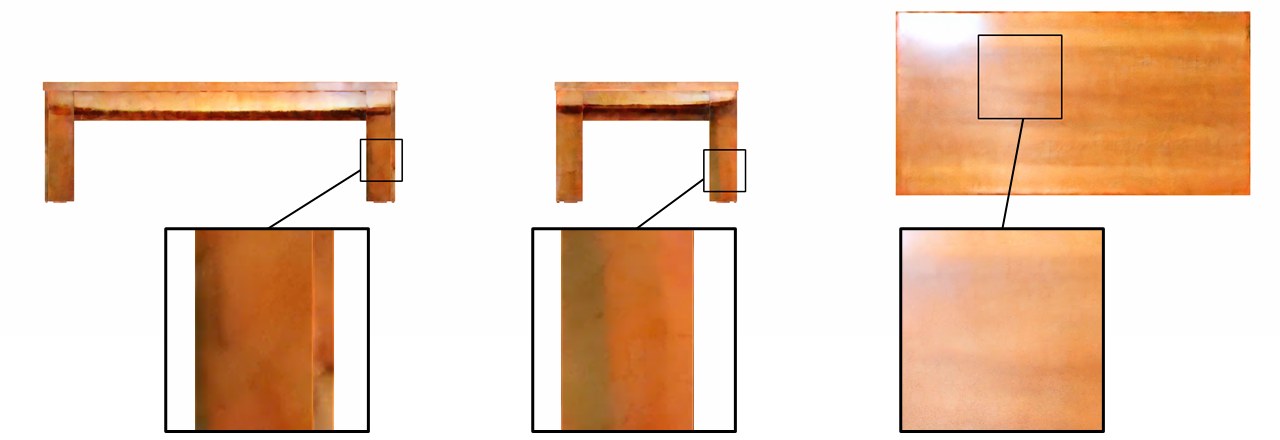}  
\end{minipage}
}
\subfigure{
\rotatebox{90}{\normalsize{~~~~TEXTure}}
\begin{minipage}[t]{0.75\columnwidth}
\centering
\includegraphics[width=0.75\columnwidth]{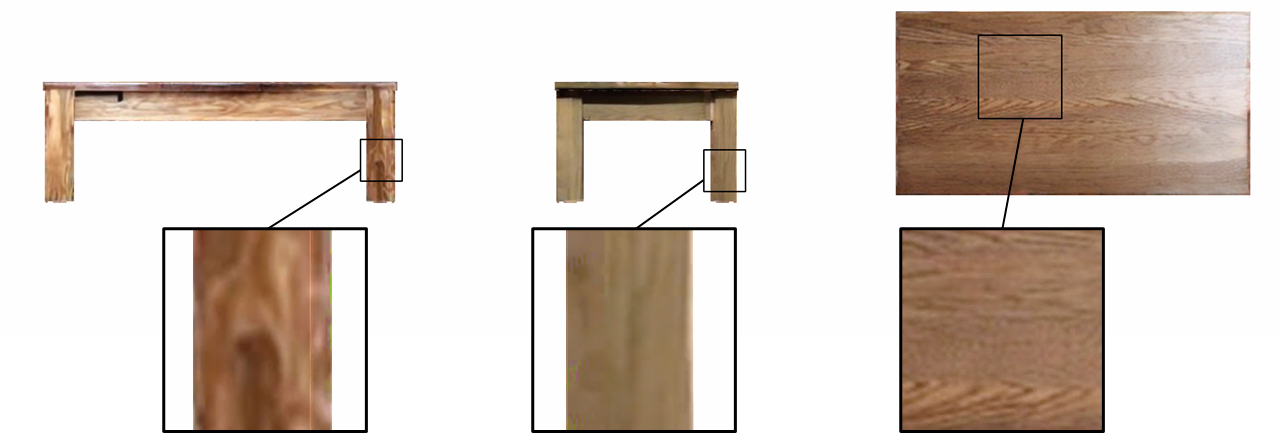}   
\end{minipage}
}
\subfigure{
\rotatebox{90}{\normalsize{~~~~~~Ours}}
\begin{minipage}[t]{0.75\columnwidth}
\centering
\includegraphics[width=0.75\columnwidth]{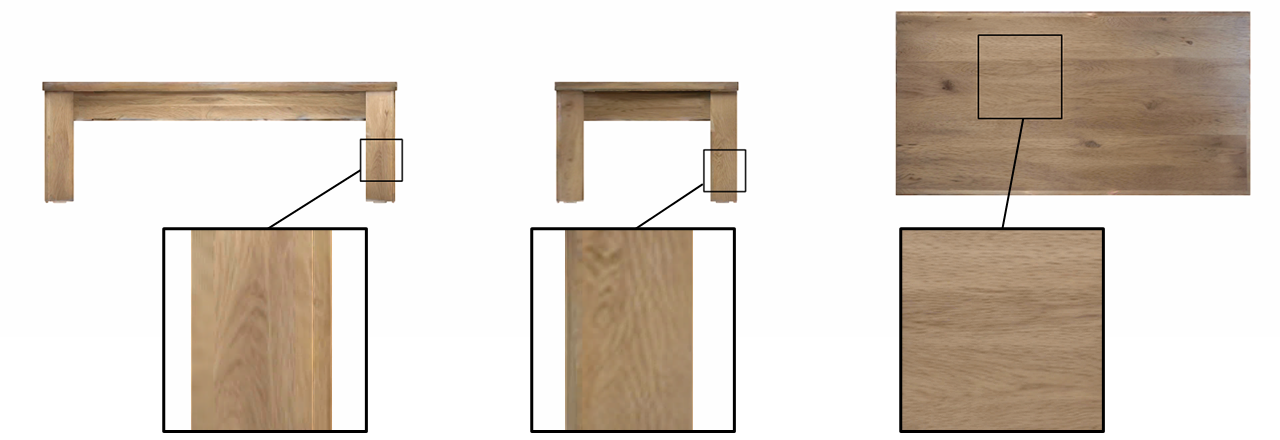}   
\end{minipage}
}
\caption{\label{fig:3}3D generated results of Latent-NeRF, TEXTure, and HCTM given the same mesh and text prompt "oak wood dining table". HCTM produces more realistic and visually appealing textures.}
\end{figure}

In the clarity experiment, we aim to evaluate the ability of our method to generate high-definition textures with clear details. We choose the text prompt "oak wood dining table" as it requires the generation of fine details and textures to accurately represent the material. As shown in Figure~\ref{fig:3}, the texture generated by Latent-NeRF is blurry and does not resemble oak at all. On the other hand, the texture generated by TEXTure is consistent with the prompt, but the details are generally unclear. In contrast, our method is able to generate clear and detailed textures, even capturing the small stripes of oak. This indicates the superiority of our method in generating high-definition textures with clear details.

\subsection{Stability}

\begin{figure}[h]
\centering
\subfigure[Latent-NeRF]{
\begin{minipage}[t]{0.3\linewidth}
\centering
\includegraphics[width=1.0\linewidth]{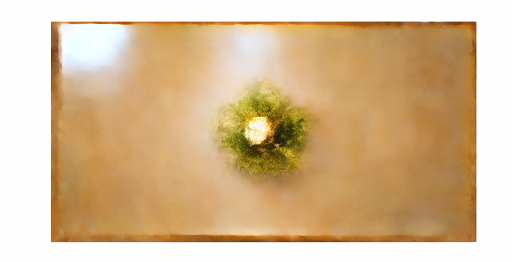}
\end{minipage}
}
\subfigure[TEXTure]{
\begin{minipage}[t]{0.3\linewidth}
\centering
\includegraphics[width=1.0\linewidth]{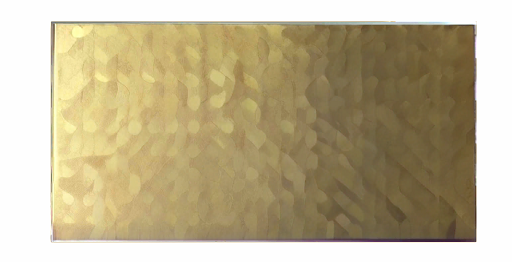}
\end{minipage}
}
\subfigure[Ours]{
\begin{minipage}[t]{0.3\linewidth}
\centering
\includegraphics[width=1.0\linewidth]{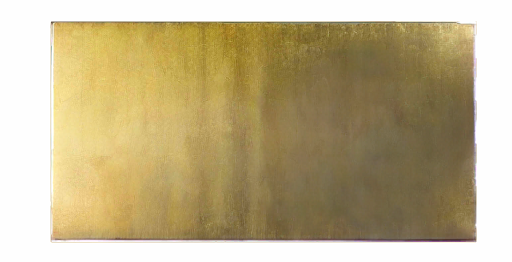}
\end{minipage}
}
\caption{\label{fig:4}3D generated results of Latent-NeRF, TEXTure, and HCTM given the same mesh and text prompt "gold dining table". HCTM produces more stable and realistic textures. }
\end{figure}

Previous methods have overly relied on Stable Diffusion, which is not omnipotent as the quality of the generated images is closely related to the prompt. Once a bad prompt is selected, the previous method will fail. For instance, "gold dining table" is a bad prompt. From Figure~\ref{fig:4}, it is evident that both Latent-NeRF and TEXTure generate textures with obvious issues, such as unreasonable patterns on the texture.

To overcome these issues, we employ reverse processes in our method. Specifically, we choose three pictures of gold from the internet, use MiDaS \cite{ranftl2020robust} to predict the depth map, and textual-inversion the prompt "gold" to S*. We then apply the LoRA technique to fine-tune the diffusion model and generate texture using the prompt "S* dining table".

As shown in Figure~\ref{fig:4}, the texture generated by our method is of high quality and is consistent with the prompt, even with a bad prompt such as "gold dining table". This result indicates that our method has good stability, which is crucial for generating high-quality textures even when the input prompt is not ideal.

\subsection{User study}

The user study is conducted with the aim to evaluate the fidelity, consistency, and quality of the generated textures on a more complex mesh and difficult prompt. The prompt chosen is "clown fish", which is a challenging task due to the intricate texture of the fish and its vibrant colors.

To perform the study, we select four methods for generating textures: Latent-NeRF, TEXTure, HCTM without multi-diffusion, and HTCM. Each respondent is asked to evaluate the generated results with respect to three aspects: (1) overall quality of the result, (2) relevance between the result and the text prompt, (3) consistency of the result.

The results is presented in Table~\ref{table1}, which shows the average scores with standard deviations for each method. It is observed that HCTM outperformed both baselines in terms of quality, relevance, and consistency. This indicates the effectiveness of our proposed method in generating high-quality and consistent textures even for complex meshes and difficult prompts. The user study also provides insights into the limitations of the baseline methods, as Latent-NeRF and TEXTure generated textures with lower fidelity and consistency compared to our method. This emphasizes the significance of incorporating our proposed techniques to achieve improved texture generation results.

\begin{figure}[h]

\centering

\subfigure[Latent-NeRF]{
\begin{minipage}[t]{0.2\linewidth}
\centering
\includegraphics[width=0.8\linewidth]{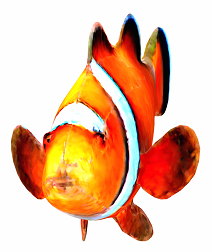}
\end{minipage}
}
\subfigure[TEXTure]{
\begin{minipage}[t]{0.2\linewidth}
\centering
\includegraphics[width=0.8\linewidth]{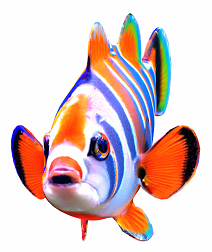}
\end{minipage}
}
\subfigure[HCTM without MD]{
\begin{minipage}[t]{0.2\linewidth}
\centering
\includegraphics[width=0.8\linewidth]{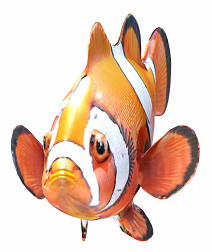}
\end{minipage}
}
\subfigure[HCTM]{
\begin{minipage}[t]{0.2\linewidth}
\centering
\includegraphics[width=0.8\linewidth]{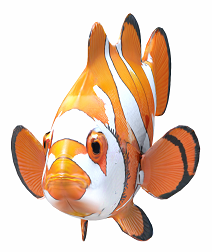}
\end{minipage}
}
\caption{\label{fig:5}3D generated results of Latent-NeRF, TEXTure, HCTM without multi-diffusion and HCTM given the same mesh and text prompt "clownfish" .}
\end{figure}

\begin{table}[h]
\caption{User study results conducted with 100 respondents. We ask respondents to rate the results on a scale of 1 to 5 with respect to the overall quality, relevance, consistence of the results. Results are averaged across all responses. }
\label{table1}
\centering
\begin{tabular}{llll}
\toprule
Methon & Quality & Relevance & Consistence \\\hline
Latent-NeRF & 1.95($\pm $0.60) & 3.33($\pm $0.85) & 3.09($\pm $0.92)  \\
TEXTure & 3.04($\pm $0.54) & 2.70($\pm $1.38) & 3.02($\pm $1.12)  \\
HCTM(without multi-diffusion) & 3.81($\pm $0.50) & 3.95($\pm $0.81) & 3.46($\pm $0.45)  \\
HCTM & \textbf{4.30}($\pm $0.60) & \textbf{4.02}($\pm $0.75) & \textbf{3.93}($\pm $0.77)  \\
\bottomrule
\end{tabular}

\end{table}

\subsection{Style transfer}
In addition to evaluating the performance of our method in terms of consistency, clarity, and stability, we also demonstrate the capability of our method in style transfer. To evaluate the performance of our method in style transfer, we conduct experiments where we transfer the style of different materials to various objects. For instance, we transfer the style of the marble dining table to eagle and chair, transfer the style of the oak wood dining table to bed and Napoleon.

Our experimental results demonstrate that our method is capable of generating high-quality style transfer textures with high fidelity, consistency, and visual appeal. The generated textures are visually similar to the source material while preserving the content of the target object. 

\begin{figure}[h]
\centering

\subfigure[marble eagle]{
\begin{minipage}[t]{0.2\linewidth}
\centering
\includegraphics[width=1.0\linewidth]{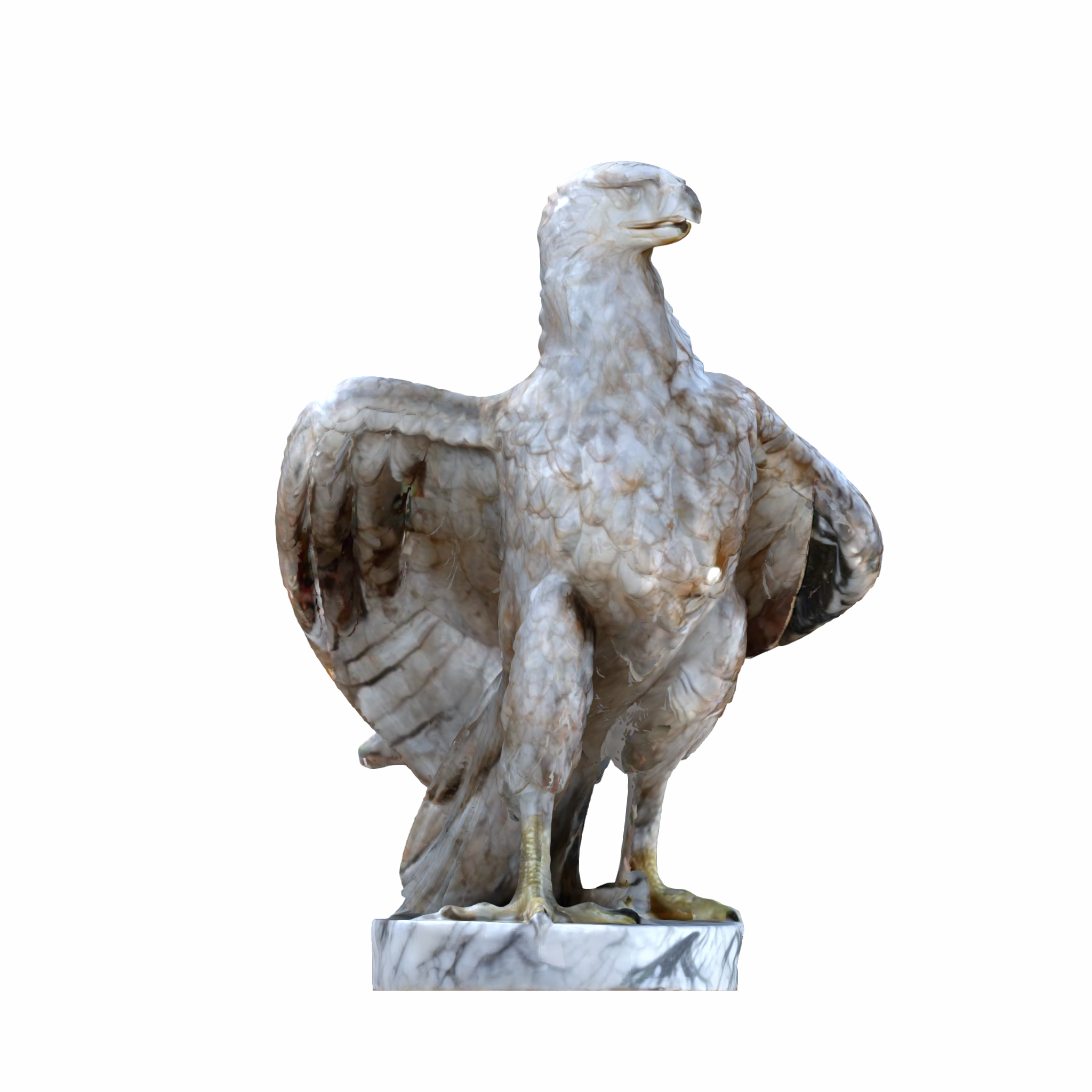}
\end{minipage}
}
\subfigure[marble chair]{
\begin{minipage}[t]{0.2\linewidth}
\centering
\includegraphics[width=1.0\linewidth]{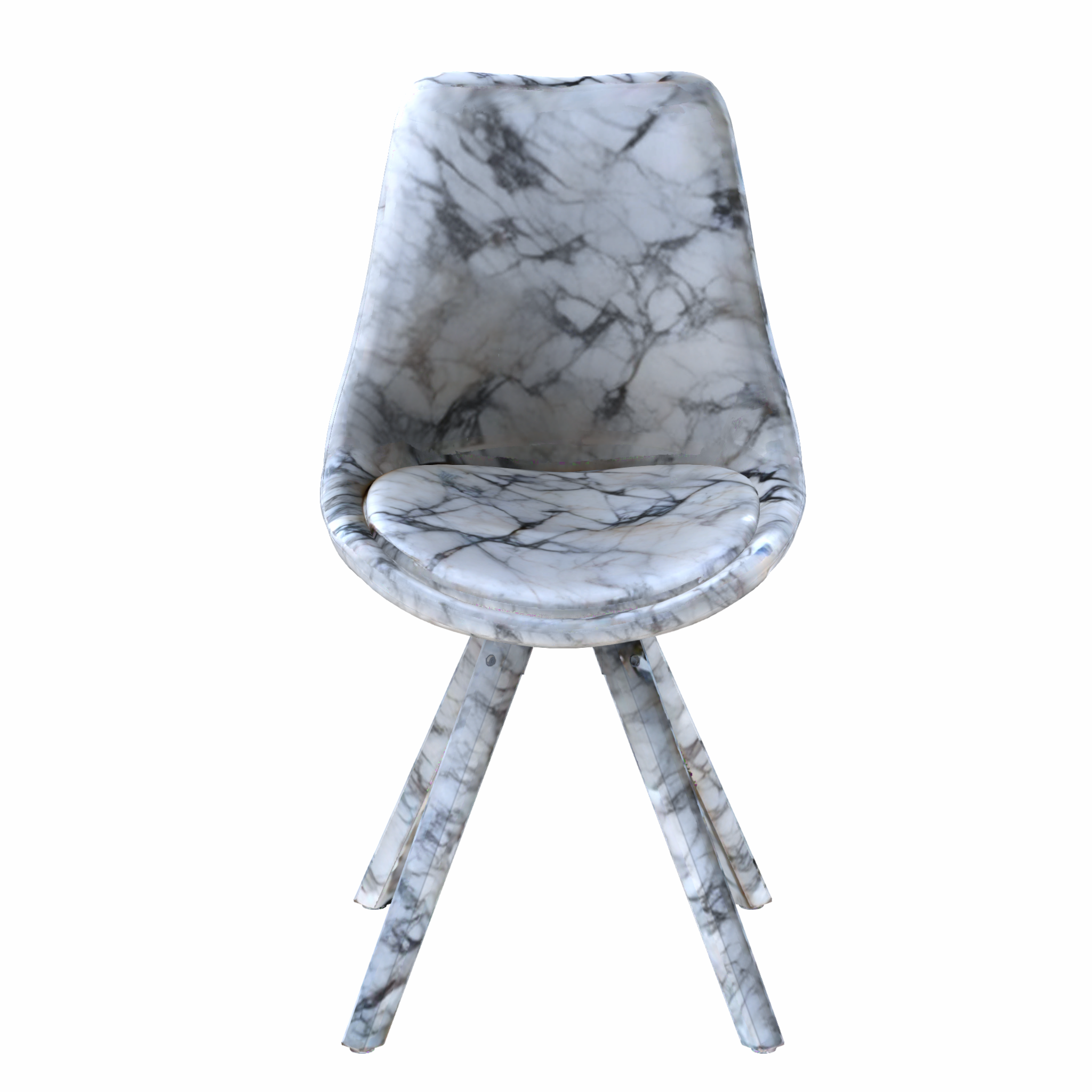}
\end{minipage}
}
\subfigure[oak wood bed]{
\begin{minipage}[t]{0.2\linewidth}
\centering
\includegraphics[width=1.0\linewidth]{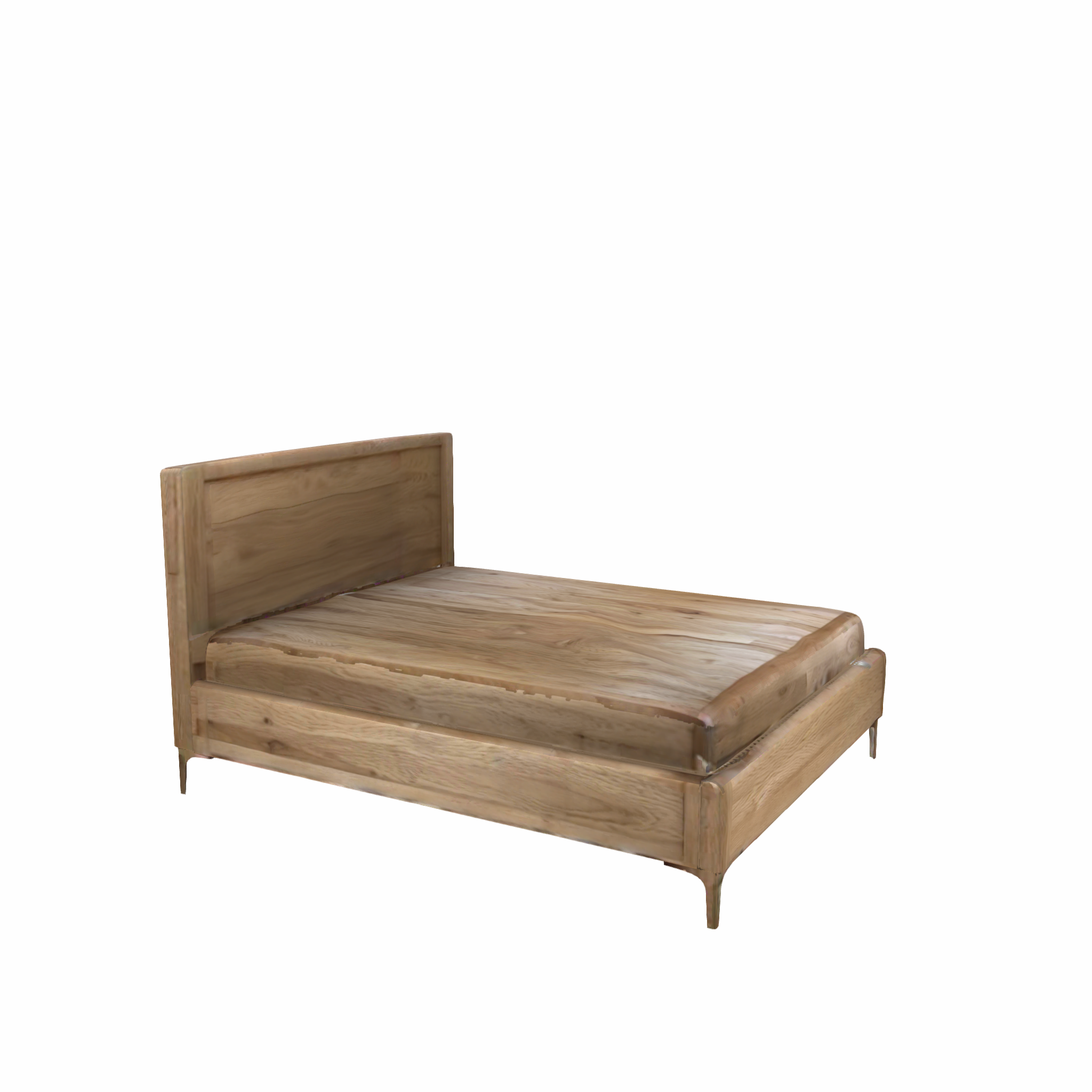}
\end{minipage}
}
\subfigure[oak wood Napoleon]{
\begin{minipage}[t]{0.2\linewidth}
\centering
\includegraphics[width=1.0\linewidth]{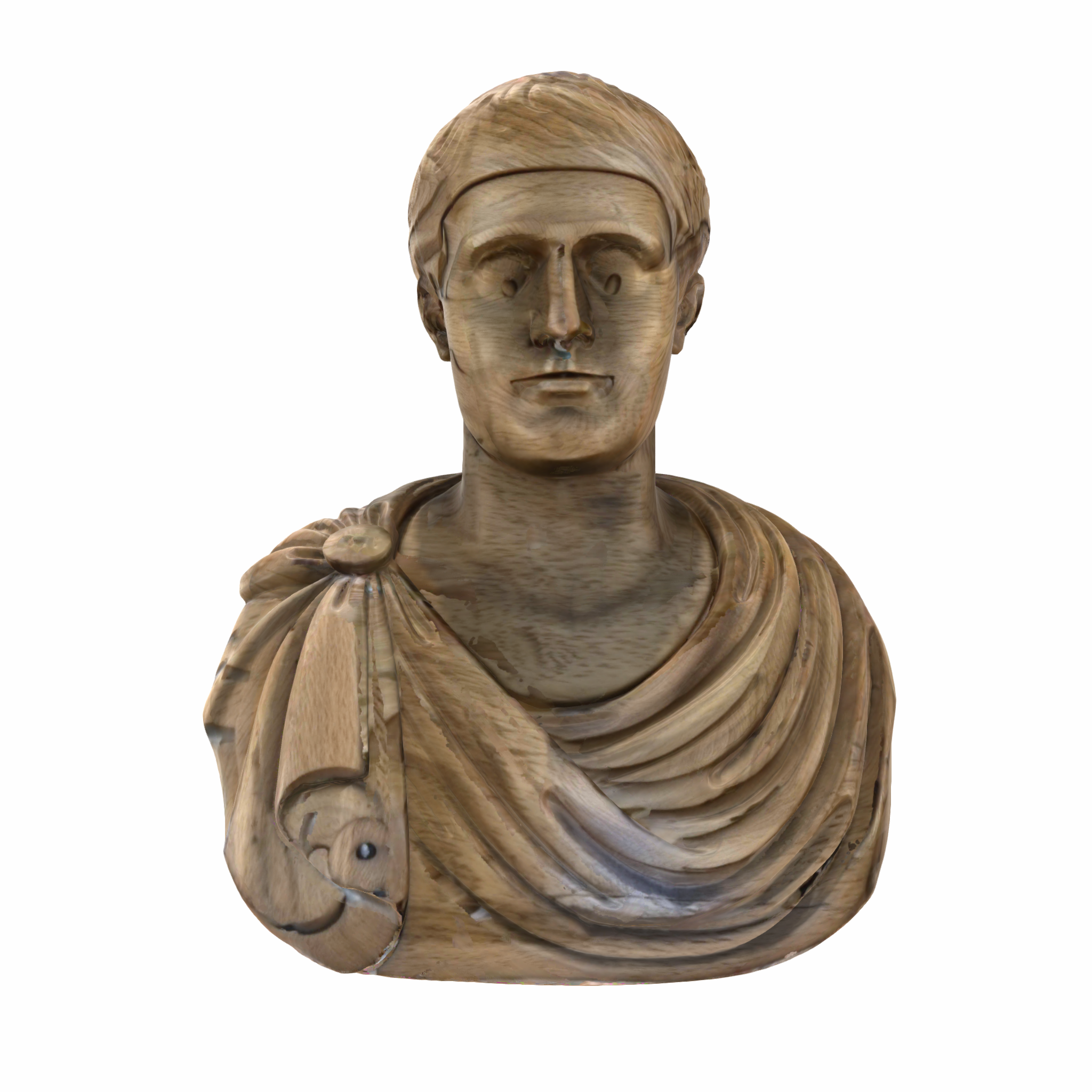}
\end{minipage}
}
\caption{\label{fig:6}One view of "marble eagle", "marble chair", "oak wood bed ", "oak wood Napoleon".}
\end{figure}
\section{Discussion, Limitations and Conclusions}
We introduce HCTM, a novel method for text-guided generating high-definition and consistent style textures. While our method has solved the problem of low-resolution and inconsistency, issues such as discontinuity, severe flare and shadows still seriously affect the resulting visuals. There have been many models for generating super-definition 2D images by generative model. However, generating high-quality 3D models is a challenging task. The first reason is that high-quality 2D image datasets are relatively easy to obtain, but the cost of making and collecting high-quality 3D models is extremely expensive. Secondly, many strategies that work in 2D space fail in 3D. For instance, the multi-diffusion strategy does not work in 3D spaces, since UV mapping changes the distribution of white noise, which is different from sliding window. Moreover, lighting has a significant influence on the visual effect of 3D models, and decoupling lighting is an extremely difficult task. Even with the ground truth image from multiple viewpoints, Nvdiffrec has made great efforts to decouple lighting but can only obtain approximate environment maps. For the generated multi-view images, which do not meet the consistency of environmental illumination, it is even more challenging to estimate the environmental illumination. Lastly, the depth-guided model is not adept at inpainting, which leads to discontinuity, and the generated images may not be consistent with the depth map.

Despite these challenges, HCTM represents a significant step forward in high-precision 3D model generation.Its ability to generate high-definition and consistent style textures from text prompt has potential applications in various fields such as gaming, virtual reality, and digital art.

{
\small
\bibliographystyle{ieee_fullname}
\bibliography{main}
}

\end{document}